\documentclass{article}

\usepackage[preprint]{neurips_2026}

\usepackage[utf8]{inputenc} 
\usepackage[T1]{fontenc}    
\usepackage{hyperref}       
\usepackage{url}            
\usepackage{booktabs}       
\usepackage{tabularx}       
\usepackage{float}          
\newcolumntype{Y}{>{\raggedright\arraybackslash}X}
\usepackage{amsfonts}       
\usepackage{amsmath}
\usepackage{bm}
\usepackage{nicefrac}       
\usepackage{microtype}      
\usepackage[dvipsnames]{xcolor}
\IfFileExists{twemojis.sty}{\usepackage{twemojis}}{\newcommand{\twemoji}[2][]{}}%
\usepackage{enumitem}
\usepackage[disable]{todonotes}
\usepackage[capitalize]{cleveref}
\usepackage{tikz}
\usetikzlibrary{positioning,arrows.meta,shapes.geometric,fit,calc}

\usepackage{listings}
\usepackage{xcolor}

\colorlet{punct}{red!60!black}
\definecolor{background}{HTML}{EEEEEE}
\definecolor{delim}{RGB}{20,105,176}
\colorlet{numb}{magenta!60!black}

\lstdefinelanguage{json}{
    basicstyle=\normalfont\ttfamily,
    numbers=left,
    numberstyle=\scriptsize,
    stepnumber=1,
    numbersep=8pt,
    showstringspaces=false,
    breaklines=true,
    frame=lines,
    backgroundcolor=\color{background},
    literate=
     *{0}{{{\color{numb}0}}}{1}
      {1}{{{\color{numb}1}}}{1}
      {2}{{{\color{numb}2}}}{1}
      {3}{{{\color{numb}3}}}{1}
      {4}{{{\color{numb}4}}}{1}
      {5}{{{\color{numb}5}}}{1}
      {6}{{{\color{numb}6}}}{1}
      {7}{{{\color{numb}7}}}{1}
      {8}{{{\color{numb}8}}}{1}
      {9}{{{\color{numb}9}}}{1}
      {:}{{{\color{punct}{:}}}}{1}
      {,}{{{\color{punct}{,}}}}{1}
      {\{}{{{\color{delim}{\{}}}}{1}
      {\}}{{{\color{delim}{\}}}}}{1}
      {[}{{{\color{delim}{[}}}}{1}
      {]}{{{\color{delim}{]}}}}{1},
}

\usepackage{amsthm}

\Crefname{equation}{Eq.}{Eqs.}
\Crefname{figure}{Fig.}{Figs.}
\Crefname{tabular}{Tab.}{Tabs.}

\newcommand\julia[1]{\todo{\textcolor{violet}{[Julia: #1]}}}
\newcommand\erlis[1]{\todo{\textcolor{violet}{[Erlis: #1]}}}
\newcommand\fabio[1]{\todo{\textcolor{violet}{[Fabio: #1]}}}

\title{\judgearena{}: A Unified Framework for Reproducible LLM-Judge Evaluation}

\newcommand{\judgearena}{\textcolor{MidnightBlue}{\textbf{JudgeArena}}}

\newcommand{\lmarenaone}{LMArena-100K}
\newcommand{\lmarenatwo}{LMArena-140K}
\newcommand{\lmarenathree}{LMArena-55K}
\newcommand{\comparia}{ComparIA}

\newcommand{\numrandomlanguages}{3}
\newcommand{\numselectedlanguages}{33}

\newcommand{\maeavgerror}{54	}
\newcommand{\amodel}{m}
\newcommand{\modelset}{\mathcal{M}}

\author{%
  Erlis Lushtaku\thanks{Equal contribution.}\, \thanks{Corresponding authors: \texttt{erlislushtaku@gmail.com}, \texttt{david.salinas.pro@gmail.com}} \\
  University of Freiburg \\
  \And
  Bora Kargi \\
  ELLIS Institute Tübingen \\
  \And
  Ali Elganzory \\
  University of Freiburg \\
  \And
  Fabio Ferreira\thanks{Work done while at University of Freiburg.} \\
  Microsoft AI \\
  \And
  Alejandro R.~Salamanca \\
  Cohere Labs \\
  \And
  Julia Kreutzer \\
  Cohere Labs \\
  \And
  David Salinas\footnotemark[1]\, \footnotemark[2] \\
  ELLIS Institute Tübingen, Prior Labs \\
}

\begin{document}

\maketitle

\begin{abstract}
LLM-as-a-judge evaluation has become a dominant paradigm for ranking language models, yet the ecosystem remains fragmented: most benchmarks ship their own code base, hardcode a specific closed-model judge, and support a single evaluation protocol. 
This fragmentation makes it difficult to study how design choices---the benchmark, the judge model, the prompt, the inference backend---affect the conclusions we draw about model quality. We introduce \judgearena{}, an open-source framework that unifies major LLM-judge benchmarks (AlpacaEval, Arena-Hard, MT-Bench, and m-Arena-Hard) under a single interface with swappable judges and comprehensive metadata logging for increased transparency in reporting and reproducibility. It enables systematic studies of judge choices, as any model accessible via vLLM, llama.cpp, or OpenRouter can serve as both candidate and judge. Furthermore, \judgearena{} ships with tuned judge configurations for open models that match or outperform closed-model judges, validated on human preference datasets in both English and multilingual settings, reducing the reliance on opaque closed models. Finally, by combining existing human annotations with LLM-judge evaluations of a target model, \judgearena{} can simulate LMArena Elo scores with high accuracy offering a practical, open, and low-cost alternative to large-scale human annotation campaigns.                 
\end{abstract}


\section{Introduction}
\label{sec:introduction}

Evaluating instruction-tuning for Large Language Models (LLMs) is challenging because generative models output free-form text in response to questions that are often open or ambiguous.
The gold standard would be human annotations, including the so-called crowd-sourcing \emph{arenas} where users annotate their preferred output to a prompt given two models\cite{chiang2024,termignon2026compariafrenchgovernmentsllm}. Since annotation costs are high, and turnaround is slow, this approach is less practical in early LLM development stages or projects with large scale evaluation needs (e.g. tracking quality across multiple languages and tasks). The alternative is to deploy an LLM judge as a human proxy~\citep{zheng2023judging}. A key advantage is that it enables automatic and cheaper benchmarks which can be run at will.

Many benchmarks relying on LLM-judges have been proposed. Among the most prominent, AlpacaEval~\cite{li2023alpacaeval,dubois2024} and Arena-Hard~\cite{li2024}, propose to annotate a candidate model on a fixed set of instructions. The answer is then compared to a baseline like a recent GPT model, and the win rates are used to construct a leaderboard. Different setups have been studied, for instance, considering multi-turn instead of single-turn in MT-Bench~\cite{zheng2023judging} or considering multiple languages in m-Arena-Hard by automatically translating instructions of Arena-Hard from English to 23 languages \cite{dang2024ayaexpansecombiningresearch}.

There is currently a clear lack of unification and reproducibility across LLM-judge benchmarks as shown in \cref{tab:framework_comparison}. For example, each benchmark ships its own independent code-base or rarely logs sufficient metadata for reproducibility. 
%
Moreover, they largely default to closed-weight judges such as GPT-4. While LLM judges exhibit self-preference bias regardless of openness \cite{panickssery2024,wataoka2025selfpreference}, closed-weight judges make this bias hard to audit and leave benchmarks vulnerable to silent updates and deprecations, rendering many evaluations stale and irreproducible\footnote{For instance, GPT-4o is the model that was used in Alpaca-Eval, Arena-Hard, m-Arena-Hard and is being retired \url{https://openai.com/index/retiring-gpt-4o-and-older-models/}}.
As a result, researchers use benchmarks coming from different packages (or write their own) which makes results sometimes incomparable~\citep{laskar-etal-2024-systematic} and strongly resembles the status of \emph{static} evaluations before the introduction of packages such as Eval Harness \cite{eval-harness} or Lighteval \cite{lighteval}.
%
This paper describes \judgearena{}, an open-source library that addresses these issues. In particular, it makes the following contributions (see \cref{tab:framework_comparison}):
    
\begin{itemize}
    \item Standardized access to the most common LLM-as-a-judge benchmarks under a unified API, with swappable judges and broad support for inference backends. 
    \item Tuned open-weight judge configurations that match or outperform closed-model judges while being cheaper and more transparent.
    \item Comprehensive metadata logging that ensures full reproducibility of every run.
    \item Extensive meta-evaluations supported by the library covering both English and \numselectedlanguages{} languages.
    \item A new LLM benchmark for estimating Elo ratings with small error across holdout models---enabling model performance estimation without participating in online arenas        
\end{itemize}


Overall, \judgearena{} unifies access to major benchmarks while exposing each meta-evaluation component for controlled study and continuous improvement.


\begin{table}[t]
\caption{Comparison of libraries supporting LLM-judges. Most support only a single benchmark, except Lighteval and Evalchemy. ``vLLM'' indicates built-in, orchestrated support for in-process local vLLM inference for both candidates and judges.}
\label{tab:framework_comparison}
\centering
\footnotesize
\newcommand{\cmark}{\textcolor{teal}{\checkmark}}
\newcommand{\xmark}{\textcolor{orange!85!black}{$\times$}}
\begin{tabular}{lcccc|ccc}
\toprule
\textbf{Framework} & \textbf{MT-Bench} & \textbf{AlpacaEval} & \textbf{Arena-Hard} & \textbf{m-Arena-Hard} & \textbf{Tuned judge} & \textbf{vLLM} & \textbf{Metadata} \\
\midrule
FastChat        & \cmark & \xmark & \xmark & \xmark & \xmark & \xmark & \xmark \\
AlpacaEval      & \xmark & \cmark & \xmark & \xmark & \xmark & \xmark & \xmark \\
Arena-Hard      & \xmark & \xmark & \cmark & \xmark & \xmark & \xmark & \xmark \\
Lighteval       & \cmark & \xmark & \xmark & \xmark & \xmark & \cmark & \cmark \\
Evalchemy       & \cmark & \cmark & \xmark & \xmark & \xmark & \xmark & \cmark \\
\textbf{\judgearena{}} & \cmark & \cmark & \cmark & \cmark & \cmark & \cmark & \cmark \\
\bottomrule
\end{tabular}
\end{table}


\section{Related Work}
\label{sec:related_work}

\paragraph{LLM-as-a-Judge Benchmarks.} Pairwise LLM-as-a-judge benchmarks have become the default cheap surrogate for human preference rankings, an alternative to crowd-sourced platforms~\cite{chiang2024} and to static benchmark mixtures~\cite{ni2024}. The pairwise paradigm was popularized
by MT-Bench and Chatbot Arena~\cite{zheng2023judging,chiang2024}, where strong LLMs were shown to agree well with crowd preferences when used as judges, despite known biases such as self-preference~\cite{panickssery2024} and verbosity. AlpacaEval~\cite{li2023alpacaeval} pairs $805$ single-turn instructions with a single GPT-4 judge scored against a fixed
reference; its length-controlled variant~\cite{dubois2024} debiases that same judge along the verbosity axis. 
Arena-Hard~\cite{li2024} curates $500$ harder English prompts auto-mined from Chatbot Arena conversations. Multilingual coverage is mostly bolted on through translation: m-Arena-Hard~\cite{dang2024ayaexpansecombiningresearch,khairi-etal-2025-life} automatically translates Arena-Hard into 23 languages and again leverages a single closed judge, while native-language preference data beyond English remain rare~\cite{termignon2026compariafrenchgovernmentsllm}
and translation-based auto-eval is known to require careful meta-evaluation~\cite{kreutzer2025dejavumultilingualllm,fu-liu-2025-reliable}. 



\paragraph{LLM Judge Frameworks.} Running LLM-judge evaluations at scale poses non-trivial engineering challenges, such as managing API calls, parsing judge outputs, and logging runs, which led benchmarks like FastChat \cite{zheng2023judging}, AlpacaEval, and Arena-Hard to provide benchmark-specific implementations tightly coupled to proprietary APIs. While users can indirectly circumvent this by manually routing requests to local OpenAI-compatible servers (e.g., standalone vLLM processes), this detached setup is error-prone and fails to capture vital inference metadata, compromising reproducibility. This architectural lock-in historically encourages reliance on closed-weight models, which are subject to silent updates and exhibit systematic self-preference \cite{panickssery2024}. Furthermore, evaluating models across different benchmarks requires managing disparate codebases with incompatible interfaces.

Generalized frameworks such as Eval Harness~\cite{eval-harness} and Lighteval~\cite{lighteval} unify diverse tasks but are optimized for static evaluation (e.g., the Open LLM Leaderboard~\cite{open-llm-leaderboard-v2}), requiring custom logic to support dynamic pairwise judges. Evalchemy~\cite{evalchemy2025} wraps existing codebases and supports vLLM for candidate generation, but its judge pipeline still relies on indirect API calls, leaving it focused on fixed proprietary judges and lacking the abstractions needed to integrate easily additional benchmarks such as Arena-Hard.

As summarized in \cref{tab:framework_comparison}, this fragmented ecosystem hinders reproducibility. \judgearena{} resolves these limitations by decoupling instruction sets from evaluator configurations. By providing a unified API with native local inference for both candidates and judges, optimized prompt templates, and rigorous metadata tracking for every generated sample, \judgearena{} seamlessly enables sovereign, open-weight evaluations with end-to-end reproducibility.



\paragraph{Meta-Evaluation and Tuning of LLM Judges.}
Meta-evaluation assesses judge reliability to determine when and for which instruction types LLM judges are suitable surrogates for human preferences.
Meta-evaluation often considers two categories of metrics: \emph{global} metrics measuring ranking correlation---Spearman rank correlation between judge rankings and Elo ratings from human arenas~\cite{li2023alpacaeval,li2024,ni2024}---and \emph{local} instruction-level agreement metrics such as accuracy or Cohen-Kappa~\cite{zheng2023judging,wang2024pandalm,zeng2024,fu-liu-2025-reliable}.
Recent work further probes judges along specific axes: hard coding and math instructions~\cite{tan2025judgebench,lai2026biasscope}, positional and verbosity biases~\cite{lai2026biasscope}, or multilingual capabilities~\cite{kreutzer2025dejavumultilingualllm,fu-liu-2025-reliable}.
Being able to measure the quality of LLM judges motivate tuning via prompt optimization~\cite{fernando2023promptbreeder} or systematic hyperparameter search over judge design choices~\cite{salinastuning}.
Despite this activity, tuned open-weight judges are rarely bundled with evaluation benchmarks; \judgearena{} directly addresses this gap by integrating meta-evaluation routines and competitive benchmarks in a single compatible framework.





\section{\judgearena{} Framework}
\label{sec:framework}

\subsection{Architecture and Design}

\begin{figure}[t]
\centering
\resizebox{\textwidth}{!}{
\definecolor{bashkw}{RGB}{0,90,160}
\definecolor{bashval}{RGB}{170,85,0}
\definecolor{termbg}{RGB}{28,28,28}
\definecolor{termfg}{RGB}{204,204,204}

\newcommand{\val}[1]{{\color{bashval}#1}}
\newcommand{\termemoji}[1]{%
  \raisebox{-0.2em}{\twemoji[height=1.15em]{#1}}%
}

\newcommand{\bashblock}{%
  \ttfamily\small
  \begin{tabular}[t]{@{}l@{}}
    {\color{bashkw}\bfseries judgearena}\\
    \quad {-}{-}task \val{alpaca-eval}\\
    \quad {-}{-}model\_A \val{LlamaCpp/./path/llama-3.2-3b-q8\_0.gguf}\\
    \quad {-}{-}model\_B \val{VLLM/utter-project/EuroLLM-9B}\\
    \quad {-}{-}judge\_model \val{OpenRouter/google/gemma-3-4b-it}\\
    \quad {-}{-}n\_instructions \val{10}\\
  \end{tabular}%
}

\newcommand{\resultblock}{%
  {%
    \setlength{\fboxsep}{6pt}%
    \colorbox{termbg}{%
      \begin{minipage}[t]{6.10cm}\vspace{0pt}%
        {\color{termfg}\ttfamily\tiny\setlength{\baselineskip}{5pt}
        \termemoji{1f4c8}\ Results Summary:\\
        \hspace*{1.55em}Total Battles: 10\\
        \hspace*{1.55em}Win Rate (A): 60.0\%\\
        \hspace*{1.55em}\termemoji{2705}\ Wins: 6\\
        \hspace*{1.55em}\termemoji{274c}\ Losses: 3\\
        \hspace*{1.55em}\termemoji{1f91d}\ Ties: 1%
        }
      \end{minipage}%
    }%
  }%
}

\begin{tikzpicture}[
    every node/.style={font=\normalsize},
    header/.style={font=\bfseries},
    boxed/.style={draw, rounded corners=3pt, inner sep=6pt, minimum height=3cm},
]

\node[anchor=north west, boxed] (cmd) at (-0.45,0) {\bashblock};

\node[anchor=west, boxed]
    (output) at ([xshift=4.50cm]cmd.east |- cmd.center)
    {\begin{tabular}[t]{@{}l@{\hspace{0.16cm}}l@{}}
      \raisebox{6pt}{\resultblock} &
      \raisebox{6pt}{\begin{minipage}[t]{2.86cm}
      \vspace{0pt}\footnotesize
      \textbf{Metadata:}
      \begin{itemize}[nosep,leftmargin=*]
        \item Package version
        \item Results JSON
        \item CLI arguments
        \item Date
        \item \ldots
      \end{itemize}
      \end{minipage}}
    \end{tabular}};

\coordinate (arrowStart) at ([xshift=0.2cm, yshift=-0.5cm]cmd.east |- cmd.center);
\coordinate (arrowEnd)   at ([xshift=-0.15cm, yshift=-0.5cm]output.west |- cmd.center);
\draw[-{Latex[length=4mm,width=3.5mm]}, line width=1.4pt]
  (arrowStart) -- node[below=2pt, font=\normalsize\itshape] {run} (arrowEnd);

\coordinate (arrowMid) at ($(arrowStart)!0.5!(arrowEnd)$);
\node[anchor=south, inner sep=0pt] (bullets) at ([xshift=-0.05cm,yshift=0.18cm]arrowMid) {%
\begin{minipage}[t]{3.95cm}\small
\vspace{0pt}
\begin{enumerate}[nosep,leftmargin=*]
  \item Pull tasks
  \item Generate completions 
  \item Generate judge annotations
  \item Store metadata and results
\end{enumerate}
\end{minipage}%
};

\end{tikzpicture}}
\caption{%
Anatomy of a \judgearena{} run. A single CLI invocation (left) selects the task, the two models and the judge as orthogonal flags; the tool pulls the task, generates or reuses cached completions, collects judge annotations, and writes the shown result (right) together with a versioned metadata bundle (package version, results JSON, CLI arguments, run date) for post-hoc reproducibility.\label{fig:architecture}}
\end{figure}
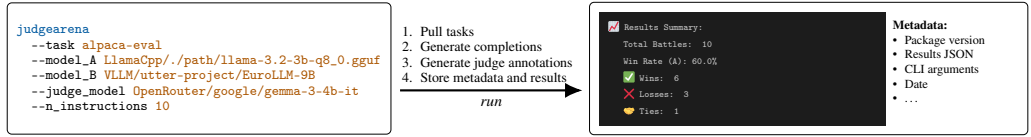

\judgearena{} is exposed through a single command-line interface, see \cref{fig:architecture}. The benchmark and the judge are selected through CLI options rather than being baked into benchmark-specific scripts. Every other aspect of a run like the truncation budgets, the chat template and the inference-engine options, is controlled through the same interface. The metadata containing all information required for reproducibility is stored in a JSON similarly to Eval Harness \cite{eval-harness}.

To perform an evaluation, instructions from the selected benchmark are loaded through a common interface and optionally subsampled to a specified number of instructions.
For each evaluated model, \judgearena{} reuses pre-shipped completions bundled by the benchmark itself when available, and otherwise generates them, persisting fresh completions through a deterministic on-disk cache.
The two completions and the judge prompt are then fed to the judge, its output is parsed into a numerical preference in $[0, 1]$, and a per-run folder of annotations, aggregate win/loss summary and metadata descriptor is written for downstream analysis.

\subsection{Inference Backends}

Model inference is factored behind a uniform backend abstraction that routes each evaluated model and each judge to the appropriate runtime.
Backends are addressed through a unified \texttt{\{Backend\}/\{ModelPath\}} naming convention; local backends run inference in-process while hosted ones are dispatched to the corresponding provider APIs via LangChain. Coverage spans vLLM (local GPU), llama.cpp (CPU and Apple Silicon), and  hosted providers such as OpenRouter or Together AI.
\judgearena{} handles both chat-templated instruction models and plain-completion pretrained base models.
Optional reasoning-token budgets are honored for thinking judges where the backend supports them, and per-request token usage is recorded across all backends.
Engine-level options such as tensor parallelism or quantization can be configured independently for the battle models and for the judge, so that a large judge can run at a different scale or precision than the evaluated models on the same GPU.

\subsection{Supported Benchmarks and Arenas}

Every benchmark follows a common instruction interface. This shared schema lets the rest of the pipeline remain agnostic to the underlying benchmark, so that adding a new benchmark only requires expressing it in the common interface rather than extending a new pipeline.

\paragraph{Benchmarks.}

The current release supports four pairwise instruction benchmarks with fixed baselines.\footnote{The choice of baselines or anchors is by itself subject to research studies~\citep{don2026mediocrity}---which we aim \judgearena{} to foster.}
AlpacaEval and Arena-Hard exposes their standard instructions together with pre-shipped completions for common baselines such as \texttt{gpt-4-1106-preview}. Arena-Hard is supported in both its v0.1 and v2.0 releases.
m-Arena-Hard brings multilingual coverage across 23 languages in two releases: v0.1 (500 prompts)~\citep{dang2024ayaexpansecombiningresearch} 
and v2.0 (498 prompts) \citep{khairi-etal-2025-life}.
\todo{David: I commenting lots since we are a bit verbose here and over 9 pages, but this content could go to the appendix.}
MT-Bench follows the FastChat protocol of 80 multi-turn questions across eight categories.


\judgearena{} preserves the original instructions of each benchmark.
\julia{This is actually not the case for m-Arena-Hard where the judge prompt is originally language-specific (which makes a big difference for spotting language confusion), see Table 9 in \citep{kreutzer2025dejavumultilingualllm}. Is this something that we can still add? We can also run an experiment that shows that in terms of correlation with human preferences. David: the instructions are preserved, but I agree that the judge prompt is changed to the default being shipped and we can reformulate this probably better. Regarding change the system prompts, I think it would be very valuable indeed, however, I am not sure we will be able to rerun all results for mAH before the deadline. Perhaps it is something we could look at post-submission? Julia: hm, I don't see how the instructions are preserved if part of the instruction is the language specification. I think it's okay if the results for submission are based on the generic prompt, but perhaps we can add an ablation on a subset that shows that modifying the prompt can have downstream changes, and therefore is important to have logged (Alejandro's PR is addressing that). David: I think it is just a problem of definition/wording, by instructions I am thinking about the set of instruction present in the HF dataset and excluding the prompt. I agree that it would be interesting to also have have per language prompts. One thing is that from Fig 3., the multilingual performance seems pretty good so the english prompt is likely not an issue but indeed would be interesting to ablate. Erlis: I did a run with translated judge prompt, and added the results in \cref{app:marena_hard_localized_prompt_ablation}, the results don't seem much different though.
}
In the runs reported here, AlpacaEval, Arena-Hard, m-Arena-Hard, and MT-Bench all use the tuned \texttt{default} preset (\cref{sec:tuned_judge}), which asks the judge for two numerical scores rather than a verdict label, so that the judge remains the sole axis of variation across these benchmarks.
As a consequence, absolute scores produced by \judgearena{} are not directly comparable to each benchmark's published leaderboard, but every run is fully reproducible from the shipped metadata.
For completeness, \cref{app:reproduction_fidelity} summarizes the original baselines, default judges and judge protocols used by the original benchmarks.





\paragraph{Arenas available.}

In addition to the benchmarking available described in \cref{tab:framework_comparison}, \judgearena{} includes also several dataset of battles from crowd source platforms which can be used for meta-analysis or to evaluate a new model. We support ComparIA \cite{termignon2026compariafrenchgovernmentsllm} as well as the three splits of LMArena splits which were previously released \cite{chiang2024} and show some of their statistics in \cref{fig:battles_stats}.
The model intersection is shown in the appendix, in particular a single model appears in all arenas which we use as a common anchor to compute Elo-ratings so that results are comparable.
The datasets have different trade-offs, \lmarenaone{} has weaker models than \lmarenatwo{} which has only recent high-performing models whereas \comparia{} and \lmarenathree{} covers all the quality spectrum. 

\begin{figure}[!h]
\center
\includegraphics[height=3.9cm]{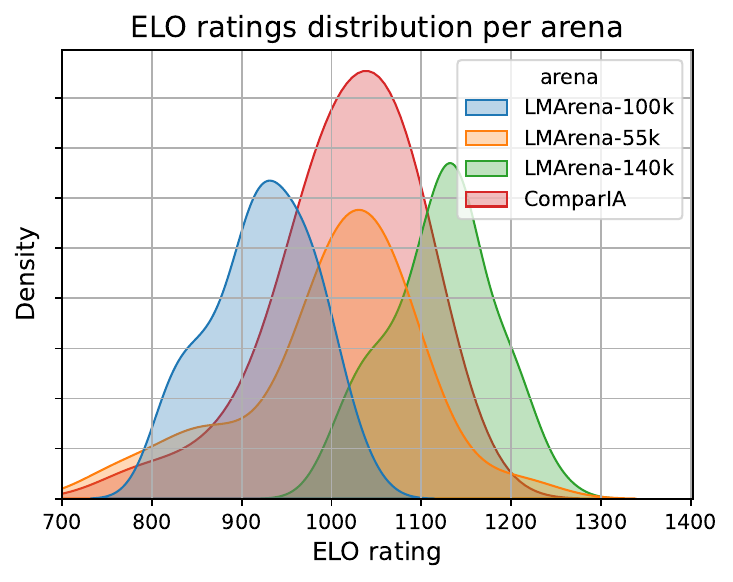}
\includegraphics[height=3.9cm]{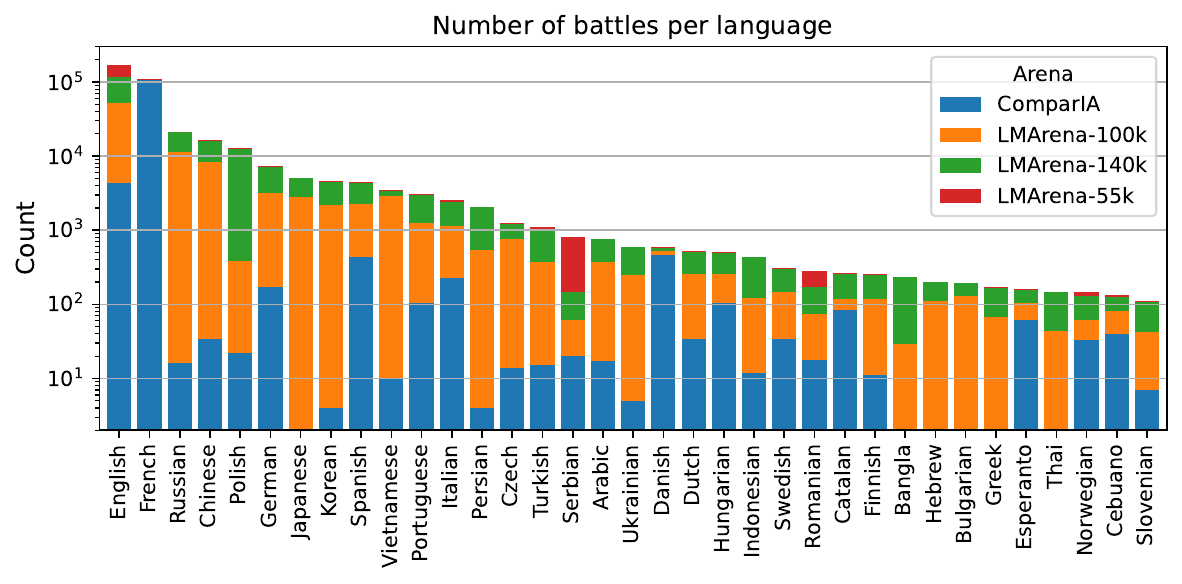}
\caption{
Elo-ratings distribution of models on different arenas (left) and number of battles per language (right).
\label{fig:battles_stats}}
\end{figure}

The number of battles per languages is also shown in \cref{fig:battles_stats} right where we show all languages having at least 100 battles. The number of battles per language varies widely for different arenas. \comparia{} has a much larger proportion of French and multilingual data than LMArena splits which all have at least half of their data as English. The combination of all those datasets is important to cover sufficiently many languages with 100 instructions which is the minimum number we set to perform meta-evaluations.


\subsection{Tuned Judge Configurations}
\label{sec:tuned_judge}

\judgearena{} operationalizes the judge-design study of \cite{salinastuning}, in which multi-objective multi-fidelity optimization was used to search over the main axes of a pairwise LLM judge: the choice of judge base model, the prompt template, the output format and scoring scheme, the decoding temperature, the use of chain-of-thought (CoT) explanations and the use of position swaps.
Rather than re-running this search for every new benchmark, \judgearena{} ships the resulting design as a default pipeline and exposes the main operational choices, including the judge model, prompt preset, explanation prompting, swap mode, and judge sampling and backend options, so that users can reuse the tuned configuration out of the box or run controlled ablations over supported axes.

The \texttt{default} preset implements the tuned protocol: the judge receives a short bias-mitigating system prompt together with a user template that asks for a structured response with two numerical scores $s_A, s_B \in [0, 10]$, one per answer.
These scores $s_A, s_B$ are then converted into a winning probability 
\begin{equation}
\mathbb{P}[A \prec B]=\sigma(\beta(s_B - s_A)) = 1 / \left( 1 + e^{-\beta (s_B - s_A)}\right)
\label{eq:judge_proba}
\end{equation}
with a sharpness factor $\beta = 0.3$.\todo{Bora: Do we need a justification for this value ß? Otherwise it feels like a magical number}
Equal scores map to a preference probability of $0.5$, while larger score differences smoothly shift the preference toward the higher-scored answer.


The remaining design axes are controlled through orthogonal configuration options.
Chain-of-thought to solicit explanation can be toggled on or off, swapping the user template for a variant that requests an explanation before the scores.
Swap evaluation runs both $A$--$B$ and $B$--$A$ orderings so that each instruction contributes two annotations rather than one, preserving per-swap information for downstream analysis; a cheaper single-pass mode is retained for large-scale sweeps.
Sampling parameters and backend-specific options are configured independently for the judge, so that the judge can run at a different scale or precision than the battle models.
Reasoning-trace management can cap the number of reasoning tokens that thinking models emit for supported backends, and visible reasoning traces can optionally be stripped from battle completions before they are sent to the judge.



\subsection{Reproducibility Infrastructure}
\fabio{if you want to shorten the paper, I think you can remove a lot here in 3.5}
Every evaluation run is treated as a self-describing artifact, following the reproducibility lessons distilled by \cite{biderman2024lessons}.
Metadata is stored containing key information such as the entry point and CLI command that produced the run, the complete run configuration (judge prompt preset, position-swap setting, truncation budgets) or the software version of main dependencies. The exact schema is detailed in \cref{app:metadata}.

In addition, a run writes a small bundle of files into a single directory whose name encodes the dataset, the two battled model sides, the judge, the position-swap setting and a timestamp: a snapshot of the full run configuration, a per-battle annotation table, an aggregate results file and a versioned metadata descriptor.
This bundle is the unit of record for each evaluation run, so that any annotation produced by \judgearena{} can be re-analyzed from its artifacts alone.

\section{Experiments}
\label{sec:experiments}

We first meta-evaluate judge quality in English and multilingual settings, then show how Elo ratings can be estimated from arenas, and finally present a case study on open-weight models across all benchmarks in the library. All scripts, completions, and annotations will be released to facilitate future research.
\todo{Bora: I think the paper above, we can use $\pm$ instead of $a^u_l$, it looks a bit crowded this way}
Throughout, we report 95\% confidence intervals via the 2.5th and 97.5th percentiles over 1{,}000 bootstrap resamples: over instructions for instruction-level metrics (Cohen-Kappa) and over models for model-level metrics (Spearman and Elo ratings).


\subsection{Meta-evaluations}
\fabio{the structure here is throwing me off a bit: the arena's paragraph doesn't belong here I think. The sub-section is called meta-evaluations but starts with two paragraphs that describe the datasets. Should this not be in section 3 where we describe benchmarks and infra, e.g., supported benchmarks? Also the two paragraphs read a bit like a dump "we have these arenas, here are the tradeoffs. We have these languages, here's the distribution". Better: connect why we are showing this, smth like "arenas are GT for the meta-evaluations that follow and we need to look at language distributions since they determine which languages we can eval at all" (perhaps have for each paragraph a paragraph-content-sentence, describing what's being described in each paragraph? would make it a more fun read).
David: I have moved the paragraph earlier, let me know if you think it fits better. I have not made it more fun to read yet, feel free to give it a try.
Fabio: much better already, thanks!}
\label{sec:exp_meta_eval}


\paragraph{English meta-evaluation.}
\fabio{suggestion: first write about the result, then the caveats for a more "appealing" read. Reorder: First paragraph describes what we found, following paragraphs describe how to read the table}
In \cref{tab:meta-analysis}, we evaluate several judges available in \judgearena{}. We consider prompted judges with both open-weights options and close-weights options with our tuned configuration (top), Alpaca-Eval and Arena-Hard prompted judges (middle) and the best fine-tuned judge reported from \cite{tan2025judgebench}. We report Cohen-Kappa, Spearman correlation and cost per annotation when selecting 100 random English instructions of \lmarenatwo{} for the top 20 models in battle counts. 

Cohen-Kappa measures the match between two annotators (higher is better) and is normalized so that values close to 0 indicates random-performance.
The values of Cohen-Kappa are competitive and outperform the Cohen-Kappa
obtained by ourselves (0.2) when re-annotating manually 200 instructions in French and English, they are far from the optimal value of $1.0$ due to the difficulty of the annotation and the noise and subjectivity inherent to the process. Looking at the Qwen3.5 series, we can see that this metric monotonically increases with the model size.  

On the other hand, Spearman rank correlation metrics are noisy and provide large confidence-intervals and much lower signal-to-noise ratio compared to Cohen-Kappa instruction level metrics as noted in \cite{salinastuning}. The Spearman rank correlation is also typically lower than observed in \cite{li2023alpacaeval,li2024} as the models selected are much closer and stronger as shown in \cref{fig:battles_stats} which is emphasized by the fact that we picked the top 20 models.

Importantly, we see that the open-weights configurations offered in \judgearena{} are competitive for both accuracy and cost as they were tuned to minimize both objectives. In particular, the tuned prompt combined with gemma-4-31b-it provides the best Cohen-Kappa agreement while being the cheapest option.


\begin{table}[t]
\caption{
Meta-evaluations for Cohen-Kappa, Spearman rank correlation and Cost (expressed in dollar-cost per 1k annotations) when evaluating English instructions.
We report results with respectively the default configuration with open-weights models and proprietary models (top), the configuration from Arena-Hard and Alpaca-Eval (medium) and fine-tuned judges (bottom).
\label{tab:meta-analysis}
}
\centering
\small
\begin{tabular}{lrrr}
\toprule
Judge & Cohen-Kappa $\uparrow$ & Spearman $\uparrow$ & Cost $\downarrow$ \\
\midrule
qwen3.5-122b-a10b & $0.15_{0.12}^{0.18}$ & $0.69_{0.29}^{0.92}$ & 4.16 \\
qwen3.5-27b & $0.14_{0.11}^{0.17}$ & $0.67_{0.28}^{0.87}$ & 4.07 \\
qwen3.5-9b & $0.12_{0.09}^{0.15}$ & $0.75_{0.41}^{0.93}$ & 0.43 \\
deepseek-v3.2 & $0.13_{0.10}^{0.16}$ & $\text{\textbf{0.78}}_{0.47}^{0.92}$ & 0.44 \\
gemma-4-31b-it & $\text{\textbf{0.17}}_{0.14}^{0.20}$ & $0.72_{0.36}^{0.91}$ & \textbf{0.25} \\
\midrule
gpt-5.4-nano & $0.11_{0.08}^{0.14}$ & $0.75_{0.39}^{0.91}$ & 0.36 \\
gemini-3.1-flash & $\text{\textbf{{0.17}}}_{0.13}^{0.20}$ & $0.73_{0.37}^{0.91}$ & 0.47 \\
\midrule
Arena-Hard + gemini-3.1-flash & $0.16_{0.12}^{0.19}$ & $0.75_{0.41}^{0.93}$ & 1.14 \\
Alpaca-Eval + gpt-5.4-nano & $0.11_{0.08}^{0.14}$ & $0.73_{0.37}^{0.91}$ & 0.38 \\
\midrule
Skywork-Critic-70B & $0.08_{0.05}^{0.11}$ & $0.48_{0.03}^{0.80}$ & 0.74 \\
\bottomrule
\end{tabular}
\end{table}

\paragraph{Multilingual meta-evaluation or assessing which language can be judged.}

To use an LLM-judge, we need to make sure that its performance is good enough for the targeted use-case. While general instructions or coding in English is well investigated, other use-cases such as multilingual instructions typically follow much less verification, which can make their results unreliable for model development and comparisons. Suboptimal choices can lead to misleading conclusions about model capabilities, as reported in \cite{dang2024ayaexpansecombiningresearch}, where results for GPT4-o vs GPT4-o-mini diverged notably. As there are large divergences in support across languages for different models, it becomes even more challenging to select a suitable judge for all. 

\begin{figure}
\center
\includegraphics[width=0.98\textwidth]{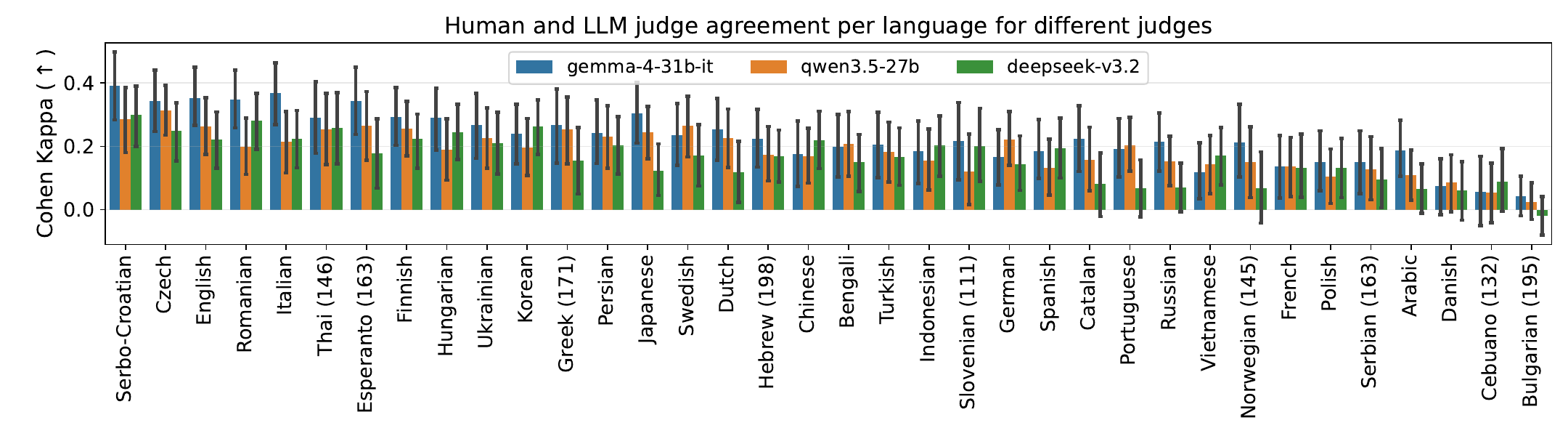}
\caption{
Cohen-Kappa per language for three different open-weight judges.
We use up to 200 random battles per language and select only languages with at least 100 battles.
The number of battles is indicated in parentheses for languages with fewer than 200 battles.
\label{fig:cohen_kappa_languages}}
\end{figure}

Any LLM judge selected to perform this task should at least clearly outperform random predictions \cite{kreutzer2025dejavumultilingualllm}.
Before evaluating and reporting results with LLM-judges on any language, we therefore evaluate whether LLM-judges outperform random predictions on all languages with more than 100 instructions in all arenas supported. In \cref{fig:cohen_kappa_languages}, we evaluate Cohen-Kappa agreement-rate for multiple languages for three strong open-weight judge configurations.
Interestingly, gemma-4-31b outperforms qwen-3.5-27b for most languages. For \numrandomlanguages{} languages, the Cohen-Kappa confidence interval overlaps zero, not discarding random-chance performance, we therefore exclude all such languages by default. We note that having higher agreement scores between LLM and human annotators for some languages does not mean that those languages are inherently easier to judge as there are other confounding effects such as annotators of a certain language looking at answer into more details or asking easier questions. However, a high agreement rate indicates that LLM judges are reliable on the given distribution.



\subsection{Elo Approximation via Hybrid Human and LLM Annotations}
\label{sec:exp_elo_approximation}

Evaluating against a single fixed baseline can cause multiple issues including non transitivity \cite{xu2025investigating} of results or self-preference \cite{panickssery2024}. To avoid this issue, \judgearena{} proposes to evaluate Elo ratings from battles previously made available in an crowd-source arena.


Elo ratings are computed given a set of battles in crowd-sourced arenas which we denote as
$(p^i, o^i_A, o^i_B, \Phi^{\text{Human}}(p^i, o^i_A, o^i_B))_{i=1}^N$
where $o^i_A, o^i_B$ denote the completions of two models $A$ and $B$ given the prompt $p^i$ and $\Phi^{\text{Human}}(p^i, o^i_A, o^i_B) \in \{-1, 1\}$ denotes the completion chosen, with $+1$ indicating that $B$ is preferred over $A$. For a given set of models $\modelset{}$, Elo ratings $\beta\in\mathbb{R}^{|\modelset|}$ are estimated via the Bradley--Terry (BT) model~\cite{bradley} such that the marginal probability that model $A$ is preferred over model $B$ on a random prompt is given by:
\begin{equation}
\mathbb{P}\!\left[\Phi^{\text{Human}}(p,o_A,o_B)=1\right] = \sigma(\beta_B - \beta_A).
\label{eq:bt}
\end{equation}
Given the set of battles, the strengths $\beta$ are estimated by maximising the log-likelihood of the observed preferences on \cref{eq:bt}, with one model strength fixed to a constant such as $1000$.
Elo scores are then obtained as $\text{Elo}(\amodel{}) = \beta_{\amodel{}} \cdot 400/\!\log 10$.

\paragraph{Estimating Elo ratings.}

\begin{figure}
\center
\includegraphics[width=0.95\textwidth]{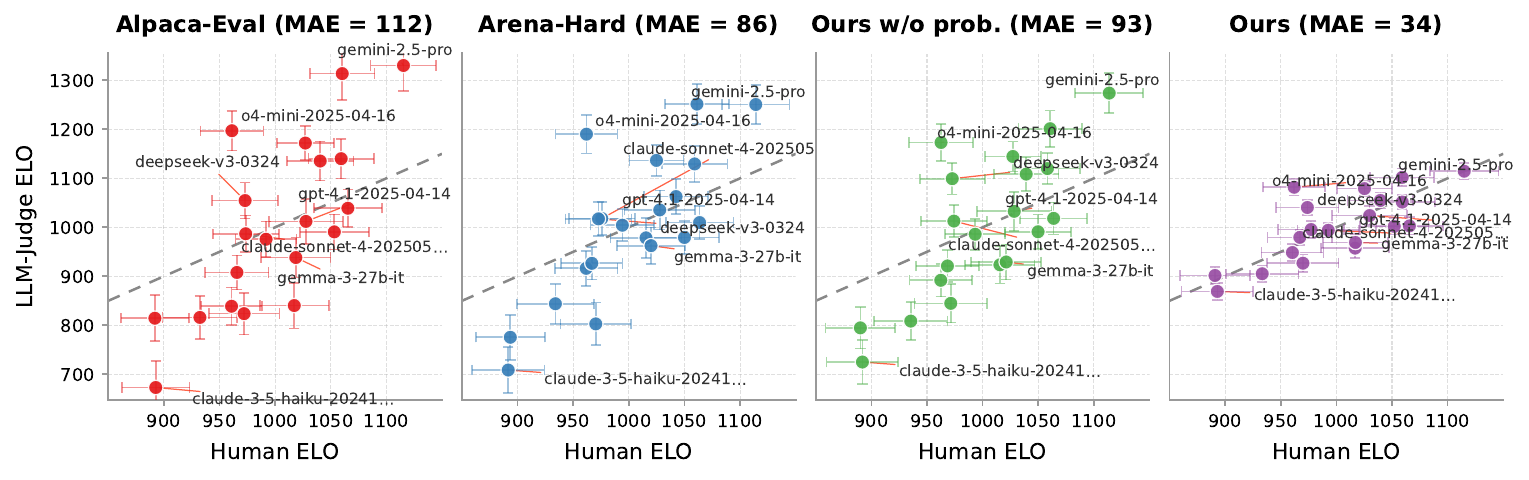}
\caption{Elo ratings estimated by LLM vs.\ human annotators for four judges.  \label{fig:scatter_elo}}
\end{figure}

To approximate Elo ratings without human annotators, we annotate 2\,000 battles from \lmarenatwo{}, 100 instructions from 20 models, with an LLM judge instead of human annotators. That is, we replace $\Phi^{\text{Human}}$ by $\Phi^{\text{LLM}}$ in \cref{eq:bt}. In \cref{fig:scatter_elo}, we show a scatter plot of Elo ratings estimated with Human and LLM annotators for each of the 20 models for four different judges. We report the average Mean Absolute Error (MAE) over models between the Elo ratings estimated by human and LLM judge. For all judges, we use \texttt{gemma-4-31b-it} as the base model to avoid confounding factors. For Alpaca-Eval and Arena-Hard, the judge produces only a hard verdict, so we fit BT with hard labels in $\{-1, 1\}$ for the losing/winning side.

Judges in \judgearena{} instead emit a continuous probability (\cref{eq:judge_proba}) which captures judge confidence: model A is more likely to beat B when scored $10$ versus $1$ than when scored $10$ versus $9$. We leverage this confidence by replacing the hard target with the predicted probability, so the BT cross-entropy is computed against continuous labels rather than discrete ones, related to richer preference targets used in reward modeling~\citep{rewardties2024, liu2024ordinal, afsharrad2026ordinal}. This up-weights battles with larger score difference (lower uncertainty) and down-weights more uncertain battles. We also report results when using only hard-labels ("Ours w/o prob.") to ablate the benefit from modeling the uncertainty obtained from LLM judgements. Accounting for uncertainty significantly lowers the MAE error. The default judge in \judgearena{} provides a faithful approximation as the MAE error on Elo ratings is $\maeavgerror{}$ on average per model. We believe this approximation can be extremely useful for model developers who cannot easily access live arenas for cost reasons.


\section{Case study}

\subsection{Evaluating Open Models Across Standard Benchmarks}
\label{sec:exp_cross_benchmark}

Having determined the quality of different judges and which languages can be judged, we conduct a case study benchmarking several open models on popular benchmarks available in \judgearena{}. We evaluate AlpacaEval, Arena-Hard (v0.1 and v2.0), MT-Bench and m-Arena-Hard~v2.0 (Arabic, Polish, Ukrainian, Chinese) for the following open models: Olmo-3-7B, Olmo-3-7B-Think, Tiny-Aya-Global, Apertus-8B, SmolLM3-3B, EuroLLM-1.7B and 9B, and Qwen3.5-9B, considering fully open models, except for Qwen3.5 and Tiny-Aya-Global which are only open-weights.
The judge is set to Gemma-4-31B-IT using the tuned \texttt{default} score-style preset across all benchmarks. 
We additionally estimate each model's LMArena Elo by combining 1{,}000 Gemma-4-judged target-model battles from a shared seeded random  LMArena-140K sample.





\begin{table}[t]
\centering
\small
\caption{
Performance of eight open-weight models on \judgearena{} benchmarks.
}
\label{tab:benchmark_results}
\setlength{\tabcolsep}{1.0pt}
\renewcommand{\arraystretch}{1.25}
\resizebox{\textwidth}{!}{%
\begin{tabular}{l cccc cccc r r}
\toprule
 & \multicolumn{4}{c}{\textbf{English / single-language}} & \multicolumn{4}{c}{\textbf{m-Arena-Hard v2.0}} & & \\
\cmidrule(lr){2-5} \cmidrule(lr){6-9}
\textbf{Model} & AlpacaEval & AH v0.1 & AH v2.0 & MT-Bench & AR & PL & UK & ZH & \textbf{LMArena Elo} & \textbf{Avg Rank} \\
\midrule
Qwen3.5-9B (think) & $0.78_{0.75}^{0.80}$ & $0.81_{0.78}^{0.84}$ & $0.50_{0.46}^{0.53}$ & $0.85_{0.80}^{0.89}$ & $0.42_{0.38}^{0.46}$ & $0.41_{0.37}^{0.45}$ & $0.42_{0.38}^{0.45}$ & $0.57_{0.53}^{0.61}$ & $1103_{1068}^{1128}$ & \textbf{1.000} \\
Olmo-3-7B-Think & $0.54_{0.51}^{0.58}$ & $0.53_{0.48}^{0.57}$ & $0.15_{0.13}^{0.18}$ & $0.71_{0.65}^{0.78}$ & $0.06_{0.04}^{0.08}$ & $0.06_{0.04}^{0.08}$ & $0.04_{0.03}^{0.06}$ & $0.23_{0.20}^{0.27}$ & $886_{845}^{912}$ & \textbf{2.111} \\
Olmo-3-7B & $0.44_{0.41}^{0.47}$ & $0.51_{0.46}^{0.55}$ & $0.13_{0.10}^{0.15}$ & $0.63_{0.56}^{0.70}$ & $0.06_{0.04}^{0.08}$ & $0.03_{0.01}^{0.04}$ & $0.03_{0.01}^{0.04}$ & $0.20_{0.17}^{0.23}$ & $854_{807}^{881}$ & \textbf{3.167} \\
SmolLM3-3B (think) & $0.28_{0.25}^{0.31}$ & $0.24_{0.20}^{0.28}$ & $0.04_{0.03}^{0.06}$ & $0.37_{0.30}^{0.45}$ & $0.05_{0.04}^{0.07}$ & $0.02_{0.01}^{0.04}$ & $0.02_{0.01}^{0.04}$ & $0.14_{0.11}^{0.17}$ & $643_{612}^{683}$ & \textbf{4.556} \\
Tiny-Aya-Global 3.35B & $0.15_{0.13}^{0.18}$ & $0.20_{0.17}^{0.23}$ & $0.02_{0.01}^{0.03}$ & $0.32_{0.26}^{0.39}$ & $0.05_{0.03}^{0.07}$ & $0.03_{0.02}^{0.05}$ & $0.04_{0.02}^{0.05}$ & $0.12_{0.10}^{0.15}$ & $600_{564}^{643}$ & \textbf{4.667} \\
Apertus-8B & $0.09_{0.07}^{0.11}$ & $0.13_{0.10}^{0.15}$ & $0.02_{0.01}^{0.03}$ & $0.17_{0.12}^{0.22}$ & $0.04_{0.03}^{0.06}$ & $0.02_{0.01}^{0.04}$ & $0.03_{0.02}^{0.04}$ & $0.13_{0.10}^{0.16}$ & $586_{531}^{643}$ & \textbf{5.833} \\
EuroLLM-9B & $0.09_{0.07}^{0.10}$ & $0.07_{0.05}^{0.09}$ & $0.01_{0.00}^{0.02}{}^{\dagger}$ & $0.18_{0.13}^{0.24}{}^{\dagger}$ & $0.03_{0.02}^{0.04}{}^{\dagger}$ & $0.02_{0.01}^{0.03}{}^{\dagger}$ & $0.02_{0.01}^{0.04}{}^{\dagger}$ & $0.10_{0.07}^{0.12}{}^{\dagger}$ & $499_{456}^{545}$ & \textbf{6.667} \\
EuroLLM-1.7B & $0.01_{0.01}^{0.02}$ & $0.01_{0.00}^{0.01}$ & $0.00_{0.00}^{0.00}{}^{\dagger}$ & $0.03_{0.01}^{0.05}{}^{\dagger}$ & $0.02_{0.01}^{0.03}{}^{\dagger}$ & $0.00_{0.00}^{0.01}{}^{\dagger}$ & $0.01_{0.00}^{0.02}{}^{\dagger}$ & $0.07_{0.05}^{0.09}{}^{\dagger}$ & $357_{290}^{415}$ & \textbf{8.000} \\
\bottomrule
\end{tabular}
}
\end{table}

Results are reported in \cref{tab:benchmark_results}. 
Qwen3.5-9B is the clear outlier in this case study, ranking first on every benchmark in the table as well as on LMArena Elo, which yields an average rank of 1.000.
The next-best models are Olmo-3-7B-Think (2.222) and Olmo-3-7B (3.056), and the gap to the rest of the table is much larger than the per-cell bootstrap intervals, so the ordering of the strongest models is stable at the instruction-sampling level.
Arena-Hard v2.0 is the most difficult English benchmark for every evaluated model, while the translated m-Arena-Hard cells show a clear drop relative to AlpacaEval, Arena-Hard v0.1, and MT-Bench.
The EuroLLM results on long-prompt cells should be interpreted with additional caution because several inputs were truncated to fit the models' 4096-token positional cap, as marked by $^{\dagger}$; the corresponding truncation statistics are reported in \cref{app:case_study_config}.

\begin{figure}[t]
\centering
\includegraphics[width=\textwidth]{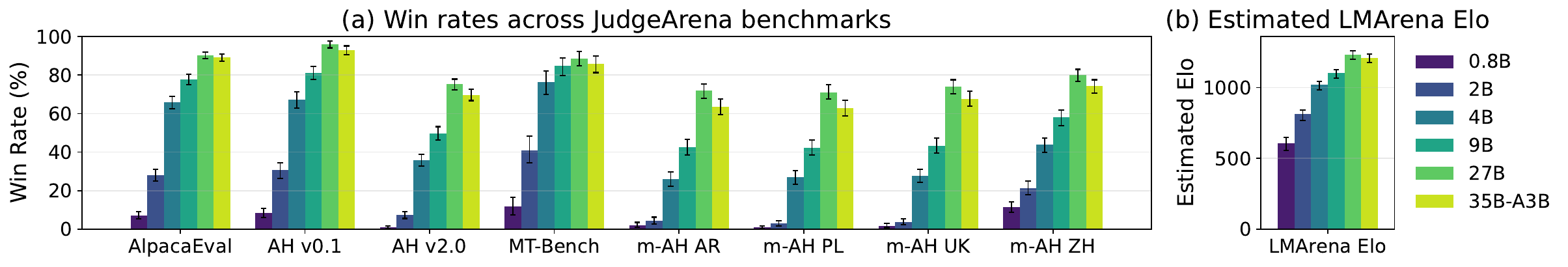}
\caption{Qwen3.5 performance scales monotonically with model size for most benchmarks. 
}
\label{fig:qwen35_monotonicity}
\end{figure}

In \Cref{fig:qwen35_monotonicity}, we study whether benchmarks show monotonic performance when scaling model parameters. We evaluate the Qwen3.5 series (0.8B--35B-A3B) with Gemma-4-31B-IT as judge; the 27B and 35B-A3B points use released FP8 variants. Except for MT-Bench and m-AH, all tasks show non-overlapping confidence intervals with monotonically improving performance from 0.8B to 27B, providing a sanity check on the discriminative power of the benchmarks in \judgearena{}.

\section{Limitations}
\label{sec:limitations}

\paragraph{Opening the blackbox.} While we allow the user to change many aspects about the judge--- the model, the temperature, the prompt---this could also lead to an undesirable situation where the judge configuration is tuned to make a particular method shine or could lead to user confusion. We believe this is alleviated by shipping a default judge configuration for every benchmark and also by reporting all metadata corresponding to a run so that a user can share the JSON corresponding to their result so that the community can reproduce their result.

\paragraph{Cross-version comparability.} It should be clear that numbers between different versions will most likely not be comparable. This is expected and similar to Eval-Harness where tasks, data from evaluations, code for generation are changed continuously, meaning any benchmark needs to be run on all methods and numbers can never be pulled from a table (unless the exact software versions are matched). \judgearena{} behaves similarly: whenever reporting a comparison such as \cref{tab:benchmark_results}, the JSON of results should be shared to enable reproducibility.

\paragraph{Limit of reproducibility.}
While we log all metadata such as software version, command-line arguments, or dataset hashes, and seed all stochastic components, some aspects are fundamentally irreproducible---among them CUDA non-deterministic execution or the use of hosted closed models whose behavior can change over time~\cite{chen2023chatgpt}. We acknowledge this limitation but still believe that having the exact code and software versions is a step up when sharing results.

\paragraph{Contamination.} 
All benchmarks available except for ComparIA are static and can be contaminated. We do not address this, as it remains an active area of research. We note that ComparIA could be used in the future to do evaluations with forward-looking data as the dataset is continuously updated.




\section{Conclusion}
\label{sec:conclusion}

We presented \judgearena{}, which lets users perform LLM-judge evaluations on multiple benchmarks including Alpaca-Eval, Arena-Hard, m-Arena-Hard, MT-Bench, ComparIA and LMArena variants.
In addition to benchmarking, the library supports performing meta-evaluations and  includes an {\em open-weight} default judge that was tuned for accuracy  and cost and will be continuously updated as newer models are released. The library supports the evaluation across \numselectedlanguages{} languages which have been selected after careful meta-evaluations. It also supports estimating Elo-ratings with small errors, which we believe will be useful for model developers who want to avoid the issue of benchmarks with fixed baselines. We hope the library will help to study, consolidate and improve LLM-judge benchmarks.

We believe the work has positive societal impact as it contributes to more transparent and reproducible LLM evaluation practices together with thorough meta-evaluation. The multilingual scope helps ensure evaluation quality is not limited to English. We encourage users to report full metadata alongside results to enable community-level scrutiny of benchmark conclusions.  

In future work, we will continue improving the accuracy of the default judge we provide. We are planning to have an automatic leaderboard to track the progress in Arenas to report simulated Elo-ratings for both English and multilingual models with publicly available completions and judgements. Finally, we are working on including rubrics which provide additional decomposition of model performance which we hope can be useful for practitioners.

\begin{ack}
This research was partially supported by the following sources: EC
under the grant No.\ 101195233 (OpenEuroLLM) and 101198470 (LLMs4EU). Views and
opinions expressed are however those of the author(s) only and do not
necessarily reflect those of the European Union. Neither the European Union nor the granting authority can be held responsible for them.

\medskip
\noindent\includegraphics[height=9mm]{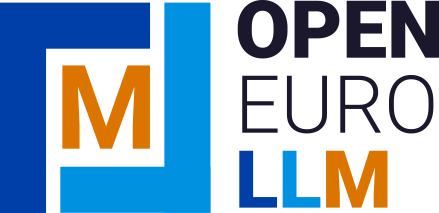}
\end{ack}

\bibliographystyle{unsrt}
\bibliography{biblio}

\clearpage

\appendix
\crefalias{section}{appendix}

\listoftodos{}

\section{Benchmark Reproduction Fidelity}
\label{app:reproduction_fidelity}
We show a summary of the differences to the original benchmarks in \cref{tab:reproduction_fidelity}.

\begin{table}[h!]
\centering
\scriptsize
\setlength{\tabcolsep}{3pt}
\caption{Reproduction fidelity of the benchmarks shipped with \judgearena{}. Instructions are loaded as released with each benchmark. The ``original default judge'' column reports the judge model used by the upstream benchmark or paper. The runs reported here use \judgearena{}'s tuned \texttt{default} preset (\cref{sec:tuned_judge}), which scores both answers on $[0, 10]$ and aggregates them via a softmax with $\beta = 0.3$ with a selected judge. Absolute numbers are therefore not directly comparable to each benchmark's published leaderboard, while every run remains reproducible from the shipped metadata.}
\label{tab:reproduction_fidelity}
\begin{tabularx}{\textwidth}{@{}l l Y Y Y@{}}
\toprule
Benchmark & Instructions & Baseline completions & Original default judge & Original judge protocol \\
\midrule
AlpacaEval        & 805                                & official \texttt{gpt-4-1106-preview}                                                                  & \texttt{gpt-4-1106-preview} (\texttt{weighted\_\allowbreak alpaca\_\allowbreak eval\_\allowbreak gpt4\_\allowbreak turbo}) & single-token \texttt{m}/\texttt{M} preference (logprob-weighted), length-controlled win rate \\
Arena-Hard v0.1   & 500                                & official \texttt{gpt-4-0314}                                                                          & \texttt{gpt-4-1106-preview} & two-pass GPT-4 judge; judge generates own reference answer before issuing 5-level verdict \\
Arena-Hard v2.0   & 750                                & official, \texttt{o3-mini-\allowbreak 2025-01-31} (hard prompt); \texttt{gemini-\allowbreak 2.0-flash-001} (creative writing) & \texttt{gpt-4.1} (config default); Gemini 2.5 (official leaderboard) & same 5-level verdict with style control (length, markdown) and per-category baselines \\
MT-Bench          & 80 multi-turn                      & \texttt{gpt-4} completions shipped with FastChat (official default is \texttt{gpt-3.5-turbo})                  & \texttt{gpt-4} for both single-answer and pairwise-baseline judgments & single-answer 1--10 grading; pairwise modes use the same GPT-4 judge \\
m-Arena-Hard v0.1 & 500 prompts $\times$ 23 languages  & Aya Expanse 8B, imported from CohereLabs/\allowbreak deja-vu-\allowbreak pairwise-\allowbreak evals           & \texttt{gpt-4o-2024-08-06} & multilingual head-to-head judging with GPT-4o \\
m-Arena-Hard v2.0 & 498 prompts $\times$ 23 languages  & Gemini 2.5 Flash, generated by \judgearena{}                                                          & \texttt{gpt-4o-2024-05-13} & same benchmark family, with GPT-4o win-rate judge \\
\bottomrule
\end{tabularx}
\end{table}

\section{Inference and Compute Details}
\label{app:compute}

All inference except for the case study battle models in \cref{sec:exp_cross_benchmark} and Skywork-Critic-Llama-3.1-70B in \cref{sec:exp_meta_eval} was
obtained through OpenRouter. For the case study battle models and Skywork-Critic-Llama-3.1-70B in meta-evaluations, model completions and judgements were generated on an L40 GPU (48\,GB VRAM) using vLLM, with the judge (Gemma-4-31B-IT) served via OpenRouter.

In the meta-evaluations results \cref{tab:meta-analysis} cost estimates are derived from OpenRouter pricing. For Skywork-Critic-Llama-3.1-70B which is not available on OpenRouter, costs are estimated using the closest equivalent in this case using the Llama-3.1-70B-Instruct price.

\section{Case-study Run Configuration and Truncation Statistics}
\label{app:case_study_config}

This appendix records the configuration and observed truncation statistics for the Gemma-4-31B-IT case study in \cref{sec:exp_cross_benchmark}.

\begin{table}[t]
\centering
\scriptsize
\caption{Run configuration for the Gemma-4-31B-IT case study. Judge-side rows
are Phase B settings; battle-model rows are Phase A cache settings that produced
the completions later judged in Phase B. The \textit{Statistic column} field
points to the corresponding observed-statistics columns in
\cref{tab:phase_a_truncations_by_model,tab:phase_b_truncations_by_model} that
each cap can influence.}
\label{tab:phase_b_caps}
\setlength{\tabcolsep}{3pt}
\renewcommand{\arraystretch}{1.15}
\begin{tabular}{>{\raggedright\arraybackslash}p{0.28\linewidth} >{\raggedright\arraybackslash}p{0.17\linewidth} >{\raggedright\arraybackslash}p{0.24\linewidth} >{\raggedright\arraybackslash}p{0.23\linewidth}}
\toprule
\textbf{Setting} & \textbf{Value} & \textbf{Statistic column} & \textbf{Short description} \\
\midrule
\multicolumn{4}{l}{\textbf{Judge-side configuration (Phase B)}} \\
\addlinespace[2pt]
\texttt{JUDGE\_MODEL} &
\texttt{OpenRouter/\allowbreak{}google/\allowbreak{}gemma-4-31b-it} &
\textit{all judge-side columns} &
OpenRouter Gemma-4 judge route. \\
\texttt{JUDGE\_PROMPT\_PRESET} &
\texttt{"default"} &
\textit{leaderboard win-rates} &
Score-style \texttt{score\_A}/\texttt{score\_B} parser. \\
\texttt{SWAP\_MODE} &
\texttt{"fixed"} &
\textit{judged battles} &
One judge call per battle; no A/B swap. \\
\texttt{PROVIDE\_EXPLANATION} &
\texttt{False} &
\textit{judge output cap hits} &
Score lines only, no rationale. \\
\texttt{TRUNCATE\_\allowbreak{}JUDGE\_\allowbreak{}INPUT\_\allowbreak{}CHARS} &
\texttt{200000} &
\textit{judge input char-cap hits} &
Judge-prompt character cap for the unified score-style judge path. \\
\texttt{STRIP\_\allowbreak{}THINKING\_\allowbreak{}BEFORE\_\allowbreak{}JUDGING} &
\texttt{True} &
\textit{thinking traces stripped} &
Removes candidate \texttt{<think>} blocks before judging. \\
\texttt{MAX\_\allowbreak{}JUDGE\_\allowbreak{}MODEL\_\allowbreak{}LEN} &
\texttt{200000} &
\textit{judge prompt length-check hits} &
Token-length check for rendered judge prompt. \\
\texttt{MAX\_\allowbreak{}OUT\_\allowbreak{}TOKENS\_\allowbreak{}JUDGE} &
\texttt{16384} &
\textit{judge output cap hits} &
Default judge \texttt{max\_tokens} budget. \\
\texttt{MT\_\allowbreak{}BENCH\_\allowbreak{}MAX\_\allowbreak{}OUT\_\allowbreak{}TOKENS\_\allowbreak{}JUDGE} &
\texttt{16384} &
\textit{judge output cap hits} &
Optional MT-Bench-specific override; set equal to the default budget in this run. \\
\texttt{JUDGE\_\allowbreak{}MAX\_\allowbreak{}CONCURRENCY} &
\texttt{32} &
\textit{none} &
Async judge-call concurrency cap. \\
\midrule
\multicolumn{4}{l}{\textbf{Battle-model generation configuration (Phase A cache)}} \\
\addlinespace[2pt]
\texttt{MAX\_\allowbreak{}MODEL\_\allowbreak{}LEN} &
\texttt{57344} &
\textit{raw/legit output cap} &
Default vLLM context window. \\
\texttt{MAX\_\allowbreak{}OUT\_\allowbreak{}TOKENS\_\allowbreak{}MODELS} &
\texttt{49152} &
\textit{raw/legit output cap} &
Default generation output budget. \\
\texttt{TRUNCATE\_\allowbreak{}ALL\_\allowbreak{}INPUT\_\allowbreak{}CHARS} &
\texttt{30000} &
\textit{prompt input~trunc.} &
Default battle-model prompt character cap. \\
\texttt{BATTLE\_\allowbreak{}THINKING\_\allowbreak{}TOKEN\_\allowbreak{}BUDGET} &
\texttt{32768} &
\textit{raw/legit thinking cap} &
Thinking-token budget for reasoning models. \\
\texttt{EUROLLM\_\allowbreak{}MAX\_\allowbreak{}MODEL\_\allowbreak{}LEN} &
\texttt{4096} &
\textit{raw/legit output cap} &
EuroLLM architectural context window. \\
\texttt{EUROLLM\_\allowbreak{}MAX\_\allowbreak{}OUT\_\allowbreak{}TOKENS\_\allowbreak{}MODELS} &
\texttt{4096} &
\textit{raw/legit output cap} &
EuroLLM generation output budget. \\
\texttt{EUROLLM\_\allowbreak{}LONG\_\allowbreak{}PROMPT\_\allowbreak{}TRUNCATION} &
\texttt{3500} &
\textit{prompt input~trunc.} &
EuroLLM char cap on long-prompt datasets. \\
\texttt{EUROLLM\_\allowbreak{}SHORT\_\allowbreak{}PROMPT\_\allowbreak{}TRUNCATION} &
\texttt{8000} &
\textit{prompt input~trunc.} &
EuroLLM char cap on short-prompt datasets. \\
\bottomrule
\end{tabular}
\end{table}

\begin{table}[t]
\centering
\scriptsize
\caption{Observed battle-model generation truncation and cap statistics from the
Phase A cache used by the Gemma-4-31B-IT judging run, aggregated by evaluated
model across all eight benchmark cells. \texttt{Prompt input~trunc.} is the
share of candidate prompts clipped by \texttt{TRUNCATE\_ALL\_INPUT\_CHARS}
before local vLLM generation. \texttt{Raw} cap rates count every generation that
ended exactly at the configured output or thinking-token budget. Qualitative inspection suggested that
many capped generations were degenerate or low-signal rather than clearly
useful continuations. The corresponding cap settings are listed in
\cref{tab:phase_b_caps}.
}
\label{tab:phase_a_truncations_by_model}
\setlength{\tabcolsep}{3.2pt}
\renewcommand{\arraystretch}{1.15}
\begin{tabular}{lrrrr}
\toprule
\textbf{Model} &
\textbf{Completions} &
\textbf{\shortstack{Prompt\\input~trunc.}} &
\textbf{\shortstack{Raw output\\cap}} &
\textbf{\shortstack{Raw thinking\\cap}} \\
\midrule
Qwen3.5-9B (think)   & 4\,207 & 0.00\,\% & 6.1\,\% & 10.7\,\% \\
Olmo-3-7B-Think      & 4\,207 & 0.00\,\% & 0.7\,\% & 4.1\,\% \\
Olmo-3-7B            & 4\,207 & 0.00\,\% & 0.1\,\% & 0.0\,\% \\
SmolLM3-3B (think)   & 4\,207 & 0.02\,\% & 10.0\,\% & 8.3\,\% \\
Tiny-Aya-Global 3.35B & 4\,207 & 0.00\,\% & 0.7\,\% & 0.0\,\% \\
Apertus-8B           & 4\,207 & 0.02\,\% & 1.9\,\% & 0.0\,\% \\
EuroLLM-9B           & 4\,207 & 4.94\,\% & 8.6\,\% & 0.0\,\% \\
EuroLLM-1.7B         & 4\,207 & 5.06\,\% & 24.1\,\% & 0.0\,\% \\
\midrule
\textbf{Total} & \textbf{33\,656} & \textbf{1.26\,\%} & \textbf{6.52\,\%} & \textbf{2.88\,\%} \\
\bottomrule
\end{tabular}
\end{table}
\erlis{we might want to report the configuration for the meta-evaluations as well. This is only for the case study.}

\begin{table}[t]
\centering
\footnotesize
\caption{Observed judge-side truncation, output-cap, and content-modification events in the
64-cell Gemma-4-31B-IT case-study runs, aggregated by evaluated model. \texttt{Judge input
char-cap hits} counts score-style judge prompts where the candidate A or B
answer was clipped before it was sent to the judge.
Thinking-trace stripping is reported separately because it is a deliberate
content normalization step, not a cap hit.
It counts stripped candidate answer fields before judging, so MT-Bench 2-turn
prompts can yield more stripping events than judged battles.
The zero values in the output-cap and prompt length-check columns document that
neither the judge generation budget nor the rendered-prompt length check was reached.}
\label{tab:phase_b_truncations_by_model}
\renewcommand{\arraystretch}{1.15}
\setlength{\tabcolsep}{2pt}
\resizebox{\textwidth}{!}{%
\begin{tabular}{lrrrrrr}
\toprule
\textbf{Model} &
\textbf{Judged battles} &
\textbf{\shortstack{Judge input\\char-cap hits}} &
\textbf{\shortstack{MT-Bench prompt\\input~trunc.}} &
\textbf{\shortstack{Thinking traces\\stripped}} &
\textbf{\shortstack{Judge output\\cap hits}} &
\textbf{\shortstack{Judge prompt\\length-check hits}} \\
\midrule
Qwen3.5-9B (think)   & 4\,207 & 16 & 0 & 4\,190 & 0 & 0 \\
Olmo-3-7B-Think & 4\,207 & 11 & 0 & 4\,284 & 0 & 0 \\
Olmo-3-7B    & 4\,207 & 11 & 0 & 21 & 0 & 0 \\
SmolLM3-3B (think)   & 4\,207 & 67 & 0 & 4\,168 & 0 & 0 \\
Tiny-Aya-Global 3.35B     & 4\,207 & 15 & 0 & 0 & 0 & 0 \\
Apertus-8B   & 4\,207 & 22 & 0 & 0 & 0 & 0 \\
EuroLLM-9B   & 4\,207 & 11 & 0 & 0 & 0 & 0 \\
EuroLLM-1.7B & 4\,207 & 11 & 0 & 0 & 0 & 0 \\
\midrule
\textbf{Total} & \textbf{33\,656} & \textbf{164} & \textbf{0} & \textbf{12\,663} & \textbf{0} & \textbf{0} \\
\bottomrule
\end{tabular}
}
\end{table}


\section{Localized m-Arena-Hard Prompt Ablation}
\label{app:marena_hard_localized_prompt_ablation}

This appendix reports the 32 m-Arena-Hard~v2.0 case-study battles from
\cref{sec:exp_cross_benchmark} when the Gemma-4-31B-IT judge uses translated
language-matched score prompts for Arabic, Polish, Ukrainian, and Chinese.
The evaluated model completions and all other judge settings are unchanged from
the main case-study runs; only the judge prompt text differs.

\begin{table}[t]
\centering
\footnotesize
\caption{Localized judge-prompt ablation on m-Arena-Hard~v2.0. The evaluated model completions and Gemma-4-31B-IT judge route match \cref{tab:benchmark_results}, but each language uses a translated score-style judge prompt instead of the English \texttt{default} prompt. \textit{Avg Rank} reports the mean rank over AR/PL/UK/ZH, with lower being better.}
\label{tab:marena_hard_localized_prompt_results}
\setlength{\tabcolsep}{4pt}
\renewcommand{\arraystretch}{1.15}
\resizebox{0.82\textwidth}{!}{%
\begin{tabular}{l rrrr r}
\toprule
 & \multicolumn{4}{c}{\textbf{Localized m-Arena-Hard v2.0 prompts}} & \\
\cmidrule(lr){2-5}
\textbf{Model} & AR & PL & UK & ZH & \textbf{Avg Rank} \\
\midrule
Qwen3.5-9B (think) & $0.42_{0.39}^{0.47}$ & $0.42_{0.38}^{0.46}$ & $0.43_{0.39}^{0.47}$ & $0.58_{0.54}^{0.62}$ & \textbf{1.000} \\
Olmo-3-7B-Think & $0.06_{0.04}^{0.09}$ & $0.06_{0.04}^{0.08}$ & $0.06_{0.04}^{0.08}$ & $0.24_{0.21}^{0.28}$ & \textbf{2.000} \\
Olmo-3-7B & $0.06_{0.04}^{0.08}$ & $0.03_{0.01}^{0.04}$ & $0.03_{0.02}^{0.05}$ & $0.22_{0.18}^{0.26}$ & \textbf{3.375} \\
Tiny-Aya-Global 3.35B & $0.04_{0.03}^{0.06}$ & $0.03_{0.02}^{0.05}$ & $0.03_{0.02}^{0.05}$ & $0.13_{0.10}^{0.15}$ & \textbf{4.375} \\
SmolLM3-3B (think) & $0.06_{0.04}^{0.08}$ & $0.02_{0.01}^{0.04}$ & $0.02_{0.01}^{0.03}$ & $0.14_{0.11}^{0.17}$ & \textbf{5.000} \\
Apertus-8B & $0.04_{0.03}^{0.06}$ & $0.02_{0.01}^{0.04}$ & $0.03_{0.02}^{0.04}$ & $0.13_{0.10}^{0.16}$ & \textbf{5.500} \\
EuroLLM-9B & $0.02_{0.01}^{0.04}{}^{\dagger}$ & $0.02_{0.01}^{0.03}{}^{\dagger}$ & $0.02_{0.01}^{0.04}{}^{\dagger}$ & $0.10_{0.07}^{0.12}{}^{\dagger}$ & \textbf{6.750} \\
EuroLLM-1.7B & $0.01_{0.00}^{0.02}{}^{\dagger}$ & $0.00_{0.00}^{0.01}{}^{\dagger}$ & $0.01_{0.00}^{0.02}{}^{\dagger}$ & $0.07_{0.05}^{0.09}{}^{\dagger}$ & \textbf{8.000} \\
\bottomrule
\end{tabular}
}
\end{table}

\begin{figure}
\center
\includegraphics[width=0.4\textwidth]{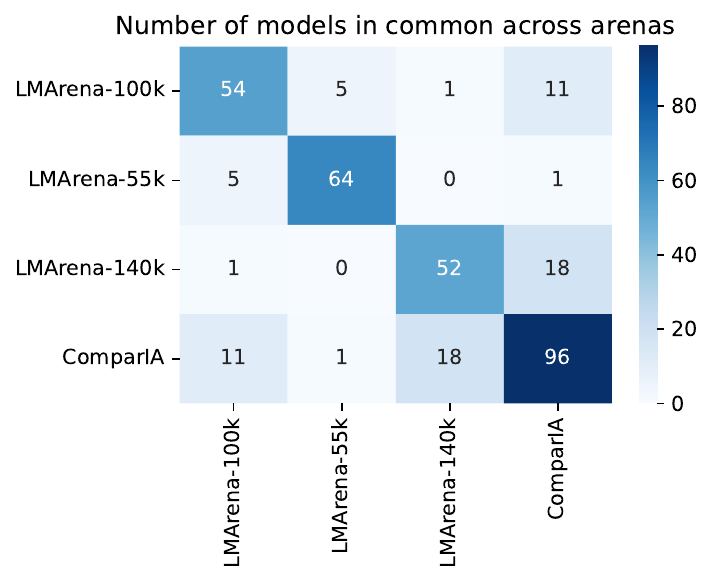}
\caption{Model intersection counts across arenas \label{fig:count_intersection}}
\end{figure}

\clearpage

\section{Metadata}
\label{app:metadata}

Every \judgearena{} run writes a versioned metadata file
(\texttt{run-metadata.\{version\}.json}) to the output directory alongside the
evaluation artifacts, see \cref{fig:metadata-example} for an example.
The version token is incremented on schema changes so that older files remain parseable.

The metadata descriptor is the central reproducibility anchor. It records the entry point that produced the run, UTC timestamps and duration, the complete run configuration (judge prompt preset, position-swap setting, truncation budgets, generation budgets and engine options), a compacted version of the results in which the per-sample preference list is replaced by its count, and counts for the logged input payloads. It also captures the execution environment: the Python version, the platform string, the installed versions of declared dependencies and the git commit hash of the \judgearena{} checkout.
When available, runtime accounting records prompt and completion token counts, while limit-event summaries report truncations and generation or reasoning caps that were hit.
To make reruns comparable without duplicating the inputs, the metadata stores cryptographic hashes of the normalized set of instruction indices and of the judge system prompt and user prompt template, so that a difference in inputs or in the judge prompt is detectable at a glance. A recursive listing of all other artifacts produced in the output directory, together with their byte sizes, closes the descriptor.


\subsection*{Field reference}

\paragraph{Schema version (\texttt{schema\_version}).}
Identifies the file layout; downstream tools use it to select the correct
parser as the format evolves.

\paragraph{Timestamps (\texttt{timestamps}).}
UTC start/end times and elapsed duration in seconds, enabling ordering of
concurrent experiments and correlation with scheduler logs.

\paragraph{Entrypoint (\texttt{entrypoint}).}
Fully-qualified Python callable used to launch the run
(e.g.\ \texttt{judgearena.generate\_and\_evaluate.main}).
Together with \texttt{git.commit} it uniquely identifies the executed code path.

\paragraph{Command (\texttt{command}).}
The exact command-line call, including all arguments, as it was run.

\paragraph{Run configuration (\texttt{run}).}
Key arguments and their resolved values at run time, including any defaults,
removing ambiguity when defaults change in later versions.

\paragraph{Results (\texttt{results}).}
Aggregate statistics: win rate, win/loss/tie/missing counts.
Per-sample annotations are saved separately as a CSV and listed in
\texttt{artifacts}.

\paragraph{Inputs (\texttt{inputs}).}
Item counts per split (instruction indices, completions from each model).
A mismatch with the annotation CSV row count indicates an interrupted or
filtered run.

\paragraph{Environment (\texttt{environment}).}
Python version and OS platform string, sufficient to flag compatibility issues
across runs.

\paragraph{Dependencies (\texttt{dependencies}).}
Pinned versions of all direct dependencies from \texttt{pyproject.toml};
uninstalled optional extras are recorded as \texttt{null}.

\paragraph{Git (\texttt{git}).}
Branch name and SHA-1 commit hash of \texttt{HEAD} at run time; omitted when
run outside a git repository.

\paragraph{Prompt hashes (\texttt{judge\_system\_prompt\_sha256},
\texttt{judge\_user\_prompt\_template\_sha256}).}
SHA-256 digests of the judge system prompt and user prompt template.
Any change to the prompt --- enabling explanation mode, switching templates,
etc.\ --- produces a different hash, making prompt variations detectable across
runs.

\paragraph{Instruction indices hash (\texttt{instruction\_indices\_sha256}).}
Order-insensitive SHA-256 digest of the sampled instruction indices, confirming
that two runs covered the same subset of prompts.

\paragraph{Artifacts (\texttt{artifacts}).}
List of all output files with relative paths and byte sizes, enabling
detection of incomplete runs.

\subsection*{Reproducibility}

The metadata captures five elements needed to reproduce a run: (1)~\texttt{git.commit}
identifies the source code; (2)~\texttt{dependencies} records the software versions;
(3)~\texttt{run} records all arguments including resolved defaults;
(4)~\texttt{command} records the exact command as typed;
(5)~prompt hashes confirm the judge prompts were unchanged.
Together, these are sufficient to fully reconstruct any evaluation.


\begin{figure}[!ht]
\begin{minipage}{\linewidth}
\begin{lstlisting}[
    %! Package Listings Error: Couldn't load requested language.                                                                        
  language=json,
  basicstyle=\ttfamily\footnotesize,
  breaklines=true,
  frame=single,
  numbers=none,
]
{
  "schema_version": "judgearena-run-metadata/v1",
  "timestamps": {
    "started_at_utc": "2026-02-27T12:51:24+00:00",
    "finished_at_utc": "2026-02-27T12:56:50+00:00",
    "duration_sec": 325.7
  },
  "entrypoint": "judgearena.generate_and_evaluate.main",
  "command": {
    "argv": ["judgearena/generate_and_evaluate.py",
             "--dataset", "alpaca-eval",
             "--model_A", "VLLM/org/ModelA-Small",
             "--model_B", "VLLM/org/ModelB-Large",
             "--judge_model", "VLLM/org/Judge-8B",
             "--n_instructions", "30", "--swap_mode", "fixed"]
  },
  "run": {
    "dataset":        "alpaca-eval",
    "model_A":        "VLLM/org/ModelA-Small",
    "model_B":        "VLLM/org/ModelB-Large",
    "judge_model":    "VLLM/org/Judge-8B",
    "swap_mode":      "fixed",
    "n_instructions": 30
  },
  "results": {
    "num_battles": 30,
    "winrate":  0.143,
    "num_wins": 3, "num_losses": 23,
    "num_ties": 2, "num_missing": 2
  },
  "inputs": {
    "instruction_index": {"count": 30},
    "completions_A":     {"count": 30},
    "completions_B":     {"count": 30}
  },
  "environment": {
    "python_version": "3.12.12",
    "platform": "Linux-x86_64-with-glibc2.28"
  },
  "dependencies": {
    "datasets": "4.5.0", "pandas": "2.3.3",
    "vllm": "0.10.2",   "transformers": "4.57.6"
  },
  "git": {
    "branch": "feat/versioned-metadata",
    "commit": "08a04362..."
  },
  "instruction_indices_sha256": "ea94c75c...0cd51",
  "judge_system_prompt_sha256": "0912ced0...239c",
  "judge_user_prompt_template_sha256": "ca30b326...ffa1",
  "artifacts": [
    { "path": "annotations.csv", "size_bytes": 155010 },
    { "path": "args.json",       "size_bytes": 483    },
    { "path": "results.json",    "size_bytes": 1041   }
  ]
}
\end{lstlisting}
\end{minipage}
\caption{Anonymised and abridged \texttt{run-metadata.v1.json} from a real
  \judgearena{} run (30 instructions, AlpacaEval dataset).
  Model identifiers replaced with placeholders; commit hash truncated.}
\label{fig:metadata-example}
\end{figure}

\section{Asset Licenses}
\label{app:licenses}

\textbf{Benchmarks.}
AlpacaEval, Arena-Hard, FastChat (MT-Bench), and m-ArenaHard (CohereLabs)
are released under the Apache~2.0 License. The Aya Expanse baseline completions for m-arena-hard-v0.1 are imported separately from \texttt{CohereLabs/deja-vu-pairwise-evals} released under
CC~BY-NC-SA~4.0.

\textbf{Human preference datasets.}
The LMArena human preference datasets used for meta-evaluation are
released under CC~BY~4.0 for user prompts; model outputs are governed
by the respective model providers' terms of use.
ComparIA conversations are released under the Etalab Open
License~2.0; the ComparIA codebase is released under Apache~2.0.

\textbf{Judge and candidate models.}
Gemma~4~31B-IT is released under Apache~2.0.
Skywork-Critic-Llama-3.1-70B is released under the Skywork Community
License.
Candidate models evaluated in the case study---Qwen3.5-9B,
OLMo-3-7B, OLMo-3-7B-Think, SmolLM3-3B, Apertus-8B, EuroLLM-9B, and
EuroLLM-1.7B---are released under Apache~2.0 while Tiny-Aya-Global is released under CC~BY-NC~4.0.
All assets are used in accordance with their respective licenses for
non-commercial research purposes.

\end{document}